\documentclass{article}

\usepackage{arxiv}

\usepackage[utf8]{inputenc} 
\usepackage[T1]{fontenc}    
\usepackage{hyperref}       
\usepackage{url}            
\usepackage{booktabs}       
\usepackage{amsfonts}       
\usepackage{nicefrac}       
\usepackage{microtype}      
\usepackage{lipsum}		
\usepackage{graphicx}
\usepackage{natbib}
\usepackage{doi}
\usepackage{amsmath}
\usepackage{graphicx}
\usepackage{subcaption}

\title{An evaluation of time series forecasting models on water consumption data: A case study of Greece}

\date{} 					

\author{ \href{https://orcid.org/0000-0001-9862-8944}{\includegraphics[scale=0.06]{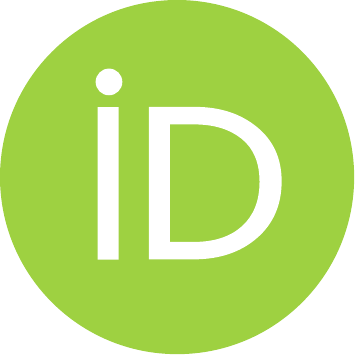}\hspace{1mm}Ioannis Kontopoulos} \\
	Department of Informatics and Telematics\\
	Harokopio University of Athens\\
	Greece \\
	\texttt{kontopoulos@hua.gr} \\
	\And
	\href{https://orcid.org/0000-0003-0514-4292}{\includegraphics[scale=0.06]{orcid.pdf}\hspace{1mm}Antonios Makris} \\
	Department of Informatics and Telematics\\
	Harokopio University of Athens\\
	Greece \\
 \texttt{amakris@hua.gr} \\
	\And
	\href{https://orcid.org/0000-0001-5183-1443}{\includegraphics[scale=0.06]{orcid.pdf}\hspace{1mm}Konstantinos Tserpes} \\
	Department of Informatics and Telematics\\
	Harokopio University of Athens\\
	Greece \\
  \texttt{tserpes@hua.gr} \\
        \And
 {
 Theodora Varvarigou} \\
	School of Electrical and Computer Engineering
       \\
	National Technical University of Athens\\
        Greece\\
        Athens Water Supply and Sewerage Company \\(EYDAP S.A.) Athens \\ 
        Greece \\
        \texttt{dora@telecom.ntua.gr} \\
}

\hypersetup{
}

\begin{document}
\maketitle

\begin{abstract}
In recent years, the increased urbanization and industrialization has led to a rising water demand and resources, thus increasing the gap between demand and supply. Proper water distribution and forecasting of water consumption are key factors in mitigating the imbalance of supply and demand by improving operations, planning and management of water resources. To this end, in this paper, several well-known forecasting algorithms are evaluated over time series, water consumption data from Greece, a country with diverse socio-economic and urbanization issues. The forecasting algorithms are evaluated on a real-world dataset provided by the Water Supply and Sewerage Company of Greece revealing key insights about each algorithm and its use.
\end{abstract}


\section{Introduction}

Nowadays, the ever-increasing urbanization and industrialization has led to a growing of water demand and a decrease in water supply and resources, thus creating a huge divergence between demand and supply. Therefore, water resources can play an important role in regional socio-economic and environmental development \citep{Setegn2015}. The effective distribution of water resources in both civil and industry life indicates the levels of urban sustainability and social inclusiveness. Proper water distribution and forecasting can act as a baseline for achieving optimal resource allocation and mitigating the gap between supply and demand, thus improving operations, planning and management.

In Greece, the recent years, the need for accurate water demand forecasting has become particularly important \citep{Bithas2006}. The systematically extraction of non-renewable ground water, the insertion of chemicals for water purification, the drought caused by climate changes in the region of the Mediterranean and the sudden rise of water demand due to the increase of refugees and migrants has created many environmental issues on the quantity and quality of the water resources as well as previously unseen socio-economic and political problems. Therefore, an accurate forecasting of water consumption can be a decisive factor for proper planning, management and optimization.

Water consumption data are seen as time series, since a measurement of water consumption levels is taken periodically (weekly, monthly, quarterly). In the literature, the problem of time series forecasting has been widely studied \citep{DEGOOIJER2006443,doi:10.1080/07474938.2010.481556,DEB2017902,SMYL202075}. Most methods for forecasting time series data include statistical models \citep{Newbold1983,Dabral2017,battineni2020forecasting} and neural networks \citep{bouktif2018optimal,xu2019hybrid,qiu2014ensemble,theodoropoulos2023graph,dingli2017financial}, which have been widely adopted due to their forecasting performance. The water consumption data of Greece pose several challenges, therefore, in this research, we evaluate the performance of several well-known forecasting models in a dataset originating from the Water Supply and Sewerage Company of Greece\footnote{\url{https://www.eydap.gr/}}. The challenges posed by the aforementioned data include the following:

\begin{itemize}
    \item A water consumption time series does not originate from one user, rather it originates from a building, a group of buildings or a block, thus aggregating the consumption levels from multiple users into one measurement or observation value and resulting in a data inhomogeneity;
    \item Observation values of the time series are not time aligned. Specifically, the month or timestamp of an observation value of one time series is not equal to the timestamp of the observation value of another time series of the same quarter and year (e.g., the observation value of the first quarter may originate from January or March). This results in a difficulty of modelling the seasonality patterns of the time series;
    \item Observation values of the time series are sparse since measurements are taken quarterly (four observation values per year);
\end{itemize}

The aforementioned challenges have driven the motivation of this research paper to perform a comparative study on forecasting models, thus identifying forecasting methodologies suitable for the data at hand.

The rest of the paper is structured as follows. Section \ref{rel_work} performs a literature review on forecasting models in various fields of study. Section \ref{models} describes the forecasting algorithms that are evaluated in this paper and Section \ref{evaluation} conducts an experimental evaluation of the algorithms. Finally, Section \ref{conslusion} concludes the merits of our work.

\section{Related Work}
\label{rel_work}

\subsection{Time series forecasting}

Gasparin et al. \citep{gasparin2019deep} provide an experimental comparison of several deep learning architectures for electric load forecasting. More specific, Feed Forward Neural Networks, Recurrent Neural Networks, Sequence To Sequence architectures and Convolutional Neural Networks are applied to the short-term load forecasting problem and evaluated against two real-world datasets. The first one considers energy consumption at an individual household while the second considers an aggregated electric demand of several consumers. The results demonstrated that simple Elmann RNNs performed comparable with gated networks such as GRU and LSTM when adopted in aggregated load forecasting. The superiority of the gated networks is observed in individual forecasting which is proof that LSTM and GRU are able to perform better regarding irregular time series. On the other hand, Sequence to Sequence models presented a moderate performance, lower than RNNs. Finally, Temporal Convolutional Neural Networks which belong to the broader category of CNNs, presented a promising performance when applied to load forecasting tasks. 

Du et al. \citep{du2018time} present a novel sequence-to-sequence deep learning framework for multivariate time series forecasting. The proposed model is a combination of a LSTM encoder and decoder component which considers the spatial-temporal dependence of multivariate time series data. The overall framework consists of two main components, a LSTM encoder which takes history multivariate time series samples and produces a fixed length vector which contains the temporal representation of the past time series data and a LSTM decoder which generates future time series based on the output vector of the first component. The proposed model is compared with a shallow learning model (SVR) and three deep learning models (RNN, LSTM and GRU). For performance evaluation, two error metrics were employed, mean square error (MSE) and root mean square error  RMSE and an air quality multivariate time series dataset. The experimental results demonstrated that proposed model achieved a satisfied accuracy.  

Bouktif et al. \citep{bouktif2018optimal} propose an optimized LSTM-RNN-based univariate model. A Genetic algorithm (GA) is employed to discover the appropriate number of time lagged features, the number of hidden layers and other important predictor variables of the proposed model. For comparison purposes, seven linear and non-linear forecasting techniques were implemented, including linear regression, ridge, regression, k-nearest neighbors, random forest, gradient boosting, ANN and extra trees regressor. Extra trees regressor was selected as the benchmark model as it presented the best performance. The performance of the forecasting techniques is evaluated using several metrics such as coefficient of variation RMSE, mean absolute Error (MAE) and RMSE. The results revealed the superiority of the LSTM-RNN forecasting model. 

Xu et al. \citep{xu2019hybrid} propose a novel hybrid model that combines a linear regression (LR) model and a deep belief network (DBN) for the prediction of time series data. The linear auto-regression model and ARIMA are employed in order to reveal the linear part of time series and then the DBN, models the nonlinear part of the time series. For the evaluation of the proposed model, three metrics were employed, MSE, normalized mean square error (NMSE) and RMSE against four different time series datasets. The results showed that the prediction accuracy of the proposed model outperforms other well-known forecasting models including LSTM, ARIMA etc. A hybrid model is able to overcome the limitations of a single model but also DBN has strong ability to extract features among layers and self-organization characteristics from observation data without prior knowledge. 

Qiu et al. \citep{qiu2014ensemble} also propose an ensemble of deep
learning belief networks (DBN) for regression and time series forecasting. The DBNs are trained using different number of epochs and then a support vector regression (SVR) aggregates the outputs of the different DBMs resulting in the final prediction. The proposed model has been compared with four benchmark methods including SVR, feedforward NN, DBN and ensemble feedforward NN. The performance evaluation is based on RMSE and Mean Absolute Percentage Error (MAPE) and the results demonstrated that the proposed model outperformed the other methods for both time series and regression datasets. To further improve the ensemble learning architecture, the same authors in \citep{qiu2017empirical} adopt the concept of ``divide and conque'' and develop a novel electricity load demand forecasting method based on Empirical Mode Decomposition (EMD) algorithm.
They propose method is an EMD algorithm and a Deep Belief Network (DBN). Ensemble learning methods are widely used as are able to obtain better forecasting performance by combining multiple learning algorithms. Specifically, the load demand data is decomposed into several intrinsic mode functions (IMF) and one residue by EMD algorithm. The DBN includes two restricted Boltzmann machines (RBMs) and one ANN is applied to each IMF including the residue. Prediction results can be aggregated by single learning machine or simply summed to obtain the final prediction. The proposed approach has been evaluated using three electricity load demand datasets and the results showed its effectiveness by comparing with nine reference methods.

Zhu et al. \citep{zhu2017deep} applied a Bayesian deep learning model for Uber time series demand prediction and uncertainty quantification. More specific, a LSTM-based auto-encoder architecture was employed with Monte Carlo dropout as both the encoder and decoder. However, unlike other models that directly use RNN to generate forecasts, the learned embedding at the end of the decoding step are fed into a Multi-layer prediction network and combined with other external features to generate the forecast. Thus, the proposed implementation provides a prediction for ``later'' values but also a level of certainty in that prediction. When a new observation falls outside a defined predictive interval, it is flagged as anomalous.

Zheng et al. \citep{zheng2017electric} develop a novel electric load
forecasting scheme based on LSTM. The proposed LSTM-based RNN scheme applies
the LSTM with peephole connections. In general, when the output gate is closed in a LSTM memory cell, the model does not know how long the memory should be retained for the model. Thus, peephole connections can be added to the LSTM memory cells as an immediate supervisor, allowing all the gates to inspect the cell states. The proposed model was compared with several forecasting methods against two different electric load time series dataset. Two error metrics were employed, MAPE and RMSE and the results demonstrated the superiority of LSTM-based forecasting method. LSTM is capable of forecasting complex univariate electric load time series with strong non-stationarity and non-seasonality.

Accurate and effective stock price prediction is considered high challenging due to the complexity and uncertainty related with the stock data.
Dingli et al. \citep{dingli2017financial} employ Convolutional Neural Networks (CNNs) for financial time series in order to forecast the next period price direction with respect to the current price. Initially, various features are identified including historic prices and technical indicators, currency exchanges, world indices etc. and then a feature selection technique statistically selects the most relevant features. The CNN achieved a moderate accuracy of 65\% when forecasting the next month price direction and 60\% for the next week price direction forecast. Ishwarappa et al. \citep{anuradha2021big} propose a hybrid model that combines a deep CNN with reinforcement LSTM model for future stock prices prediction based on big data. The original data are pre-processed using HDFS and MapReduce for extracting particular information of the stock market details. For the evaluation of the proposed model, several metrics are employed such as MAPE, Average relative variance (ARV), Prediction of change in direction (POCID) and Coefficient of Determination $R^{2}$ against four datasets of stock markets namely NASDAQ, BSE, TAIEX, and FTSE. The results demonstrated that the proposed model performed better in comparison with other techniques including FLANN-GA, MLP, LSTM and BPNN.

Sagheer et al. \citep{sagheer2019time} propose a Deep LSTM (DLSTM) architecture 
for time series forecasting. The proposed model is an
extension of the original LSTM model. It includes multiple LSTM layers such that each layer contains multiple cells. Several LSTM blocks are stacked and connected in a deep recurrent network fashion in order to combine the advantages of a single LSTM layer. Each LSTM layer operates at different time scale and passes a certain part of the task to the next layer, until finally the last layer generates the output. A Genetic Algorithm is employed for optimal selection of the proposed model hyper-parameters. The proposed model was compared with different reference models including statistical methods, machine learning, and hybrid methods. Two different error metrics are used, RMSE and root mean square percentage error (RMSPE) and two datasets which contain raw production data of two actual oilfields. The empirical results showed that the proposed DLSTM model outperforms other standard approaches and can describe efficiently the nonlinear relationship between the system inputs and outputs.

Shastri et al. \citep{shastri2020time} propose three deep learning based models to predict COVID-19 confirmed and death cases for India and USA. The deep learning models employed are LSTM variants i.e. Stacked LSTM, Bidirectional LSTM and Convolutional LSTM. Stacked LSTM also known as multi-layer fully connected structure is comprised of multiple LSTM layers forming a stack-like architecture. Bidirectional LSTM is able to process the information in both directions with different hidden layers as forward layers and backward layers and Convolutional LSTM is an extension of simple LSTM model which is able to read 2D spatial-temporal data. Four datasets of India and USA are employed which contain COVID-19 confirmed and death cases. The results showed that Convolutional LSTM outperformed the other models as it presented the minimum MAPE.

Rajagukguk et al. \citep{rajagukguk2020review} present four deep learning models for solar irradiance and photovoltaic (PV) power prediction. More specific, three standalone models were selected namely RNN, LSTM and GRU and a hybrid which uses CNN layers for feature extraction on input data and combined with LSTM to support sequence prediction (CNN-LSTM). Several evaluation metrics are utilized such as MAE, MAPE, Mean bias error (MBE), Relative Mean bias error (rMBE), rRMSE, RMSE and Forecasting skill. The results showed that the hybrid model (CNN–LSTM) outperformed the three standalone models in predicting solar irradiance. In addition, Sorkun et al. \citep{sorkun2017time} compare several deep learning models for solar irradiation data time series forecasting. LSTM and GRU models are compared against typical RNN and Naive models using hourly global horizontal solar radiation data. As the results suggested, LSTM and GRU models outperformed the conventional simple RNN architectures. 

Salman et al. \citep{salman2015weather} investigate several deep learning techniques for weather forecasting. Weather forecasting aims to develop a robust weather prediction model which exploits hidden structural patterns in the massive volume weather dataset. The prediction performance of RNN, Conditional Restricted Boltzmann Machine (CRBM) and Convolutional Network (CN) models is compared using two different datasets. The forecasting accuracy of each model is evaluated using Frobenius norm. Furthermore, the results showed that RNN model which is implemented using heuristically optimization method, is able to provide an adequate accuracy in rainfall prediction. 

Shen et al. \citep{shen2020novel} introduce a novel time series forecasting model, named SeriesNet, which can fully learn data features in different interval lengths. The proposed model is in fact a combination of two networks, a LSTM which aims to learn holistic features and to reduce dimensionality of multi-conditional data and a dilated causal convolution network which aims to learn different time interval. Furthermore, to improve generalization and in order to improve accuracy, residual learning and batch normalization are employed. The performance evaluation is conducted using three open series datasets and three different error metrics were employed, RMSE, MAE and $R^{2}$. The results showed that the proposed model achieved a higher forecasting accuracy when compared with other deep learning models including LSTM, UFCNN, ANN and SVM.


\section{Time series forecasting models}
\label{models}

In this section, the time series forecasting algorithms are described that are used in the experimental evaluation of the water consumption use case.

\subsection{Baseline forecasting model}

Before applying any complex algorithms for the forecasting of the water consumption time series, an algorithm is needed that can act as a baseline. Specifically, we needed an algorithm that uses simple rules for the forecasting of the next consumption value in the time series on which we can improve upon, thus acting as an evaluation measure. To this end, an algorithm was developed that follows the simplistic assumption that the next consumption or observation value is equal to the mean of the two previous observation values of the same season:

\begin{equation}\label{eq:baseline}
    y_{forecast} = \frac{y_{i-s} + y_{i-s \times 2}}{2}
\end{equation}

where $y$ is the observation value, $i$ is an integer index denoting the $i^{th}$ element of the time series and $s$ is the number of steps required to get the observation value of the previous or the next year. For instance, in a time series where consumptions are reported once every quarter (quarterly), $s$ would be equal to $4$. According to Equation \ref{eq:baseline}, the sum of the observation value of the previous year ($i-s$) and the observation value two years ago of the same season ($i - s \times 2$) is divided by two to give an average observation value which is then used as a forecast.

\subsection{SARIMA}

SARIMA stands for Seasonal Autoregressive Integrated Moving Average and is a well-known algorithm in the field of time series forecasting. SARIMA is an extension to the ARIMA model \citep{Newbold1983} able to support the modeling of the seasonal component in a seasonal series -- a time series with a repeating cycle. Both algorithms are widely used as standard methods for univariate time series data forecasting. In the SARIMA model, seasonal differencing is applied to remove non-stationarity from the series. Non-stationarity refers to the fact that the unconditional joint probability distribution of the mean does not change when shifted in time. A first order seasonal difference is the difference between an observation and the corresponding observation from the previous year and is calculated as $z_{i} = y_i - y_{i-s}$. Similar to Equation \ref{eq:baseline}, $s$ is equal to $4$ for quarterly time series data and $s=12$ for monthly time series data, while $i$ is an integer index. The SARIMA model is often referred as $SARIMA(p,d,q)\times(P,D,Q)^s$, where $p$ is the trend autoregression order, $d$ is the trend difference order, $q$ is the trend moving average order, and $P$, $D$, $Q$ is their seasonal counterpart respectively. The SARIMA model has been widely used in the literature for time series forecasting \citep{Dabral2017,XU2019169,6676239} and is therefore a suitable method to be evaluated for the task at hand. More details on SARIMA can be found here \citep{ForecastingPrinciples}. For the evaluation, a pre-processing step was performed to find the best parameters of the model for each time series.

\subsection{FB-Prophet}
\label{fbprophet}

FB-Prophet is an open-source framework designed by Facebook’s Core Data Science team. It is employed for time series data forecasting, based on an additive regression model.
FB-Prophet is optimized for the business forecast tasks at Facebook, for example time, daily, weekly observations of history, within a year, large outliers, missing observation and trends that are non-linear growth curves \citep{yenidougan2018bitcoin}. It performs more effectively with time series that have strong seasonal effects and several seasons of historical data. A well-derived FB-Prophet model can also used to detect anomalies and fill gaps in missing values \citep{battineni2020forecasting}. For the evaluation, a pre-processing step was performed to find the best parameters of the model for each time series.

\subsection{Long-Short Term Memory neural networks}
\label{LSTM}

Long Short-Term Memory (LSTM) was firstly introduced by
Hochreiter and Schmidhuber in \citep{hochreiter1997long} to overcome the vanishing and exploding gradients problems occurring in standard RNNs, when dealing with long term dependencies. In the standard RNN, the overall neural network is a chain of repeating modules formed as a series of simple hidden networks. In contrast, the hidden layers of LSTM introduce the concepts of gate and memory cell in each hidden layer. 
The gates in an LSTM unit enables it to preserve a more constant error that can be back-propagated through time \citep{patterson2017deep}. 

\begin{figure}
\centering
  \includegraphics[width=.4\linewidth]{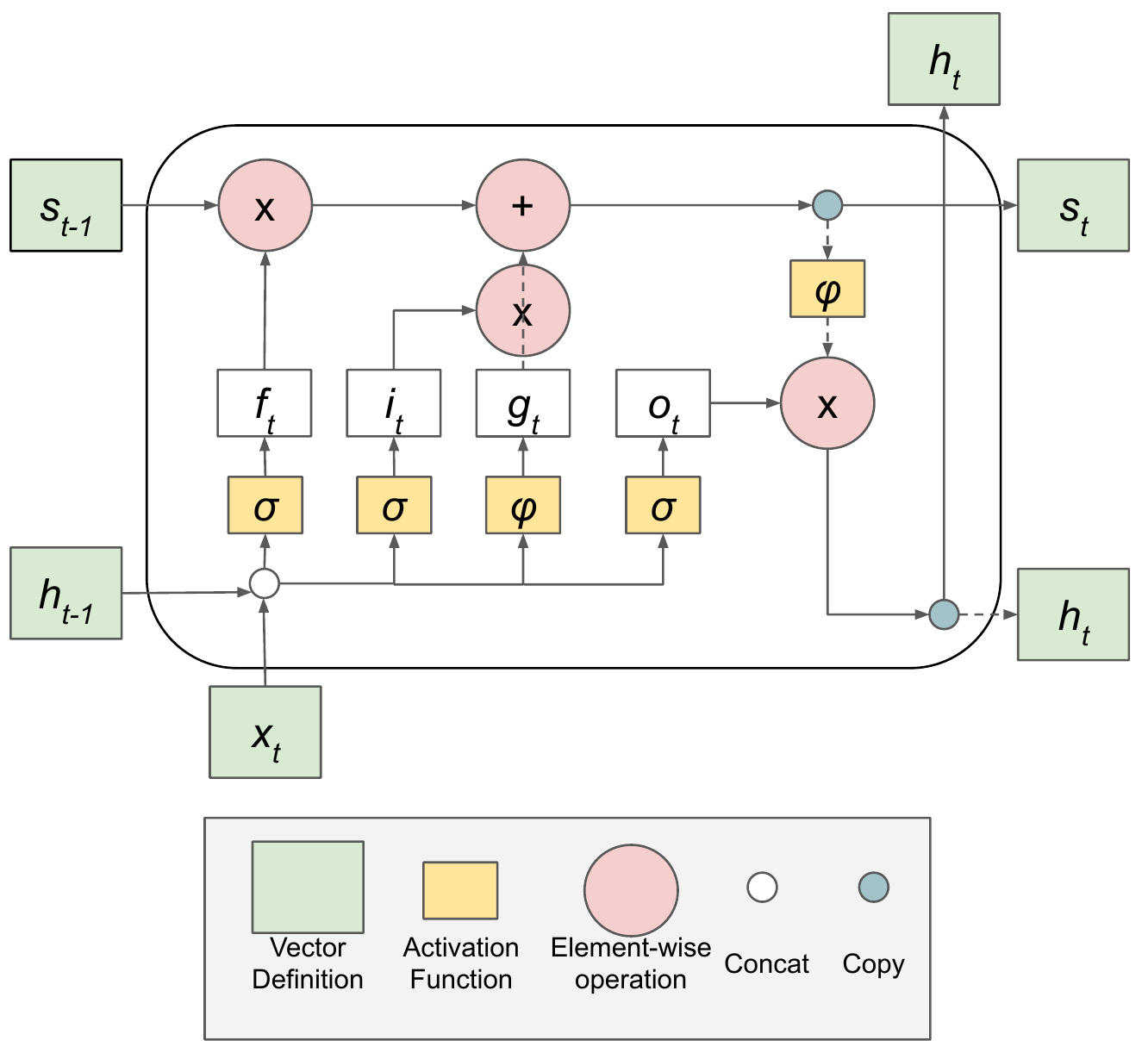}
  \caption{Structure of a LSTM block}
  \label{fig:lstm_architecure}
\end{figure}

In order to establish temporal connections, LSTM maintains an internal memory cell state throughout the whole life cycle.
The memory cell state $s_{t - 1}$ interacts with the intermediate
output $h_{t - 1}$ and the subsequent input $x_{t}$ to determine which elements of the internal state vector should be updated, maintained or forgeted based on the outputs of the previous time
step and the inputs of the present time step.
In addition to the internal state, the LSTM structure also defines an input node $g_{t}$, an input gate $i_{t}$ which controls the entry of the activations to the memory cell, a forget gate $f_{t}$ which is in charge of reset the memory cells by forgetting the past input data and an output gate $o_{t}$ which determines the value of the next hidden state. The LSTM block structure at a single time step is illustrated in Figure \ref{fig:lstm_architecure}. The neural computations in an LSTM structure are:

\begin{equation}
f_{t} =  \sigma (W_{fx}x_{t} + W_{fh}h_{t-1} + b_{f})
\end{equation}
\begin{equation}
i_{t} = \sigma (W_{ix}x_{t} + W_{ih}h_{t - 1} + b_{i}) 
\end{equation}
\begin{equation}
g_{t} = \varphi  (W_{gx}x_{t} + W_{gh}h_{t - 1} + b_{g}) 
\end{equation}
\begin{equation}
o_{t} = \sigma  (W_{ox}x_{t} + W_{oh}h_{t - 1} + b_{o}) 
\end{equation}
\begin{equation}
s_{t} = g_{t} \odot i_{t} + s_{t - 1} \odot f_{t}
\end{equation}
\begin{equation}
h_{t} = \varphi(s_{t}) \odot o_{t}
\end{equation}

where $W_{fx}$, $W_{fh}$, $W_{ix}$, $W_{ih}$, $W_{gx}$,$W_{gh}$, $W_{ox}$, $W_{oh}$ are weight matrices for the corresponding inputs of the network activation functions; $x_{t}$ is the input vector; $h_{t - 1}$ is the previous hidden state; $b_{f}$, $b_{i}$, $b_{g}$ and $b_{o}$ are the corresponding bias vectors; $\odot$ represents the element-wise multiplication; $\sigma$ is generally a sigmoid activation and $\varphi$ represents a non-linear function.

\subsection{Gated Recurrent Units neural networks}
\label{GRUs}

\begin{figure}
\centering
  \includegraphics[width=.4\linewidth]{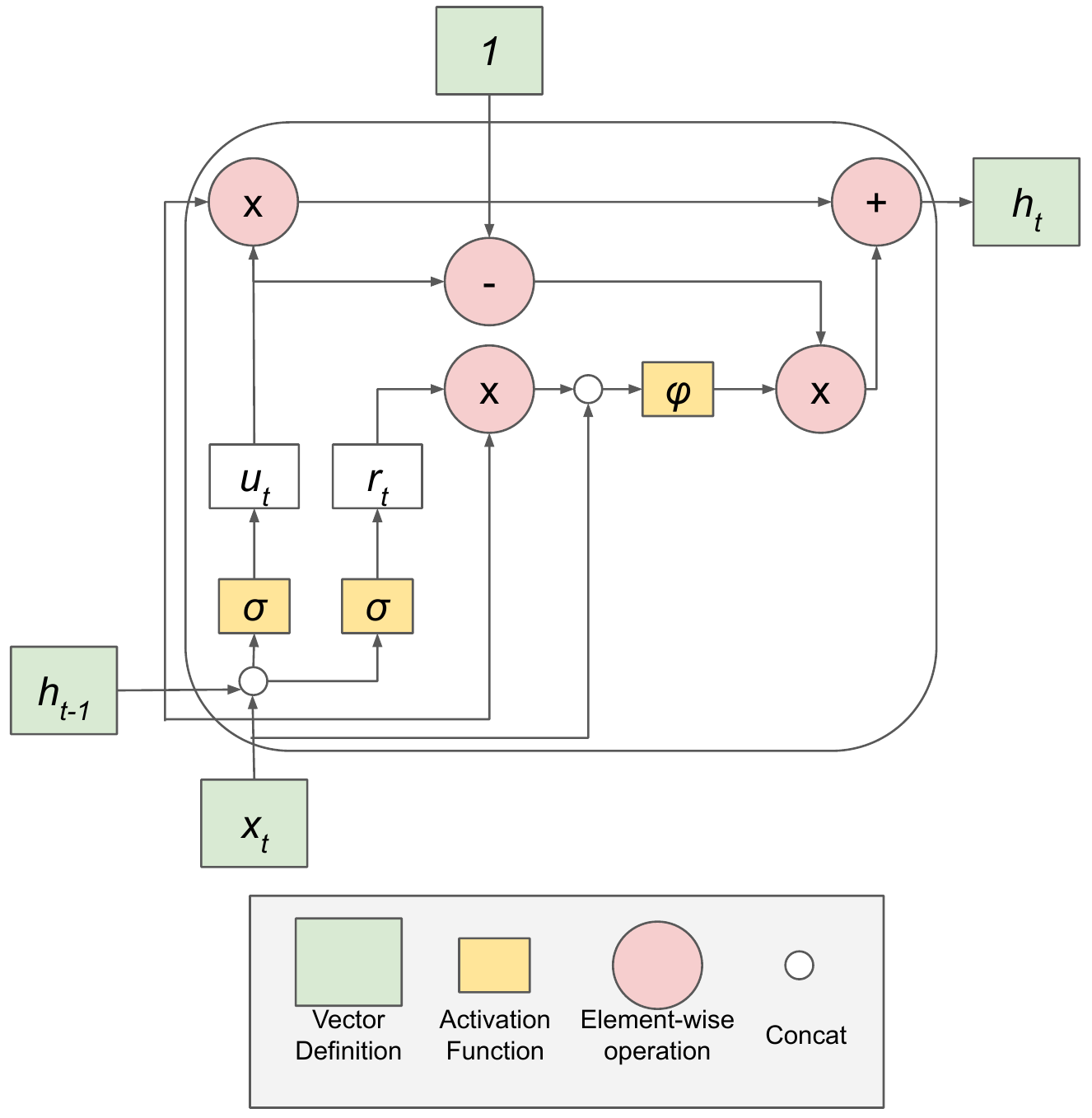}
  \caption{Structure of a GRU memory block}
  \label{fig:gru_architecure}
\end{figure}

GRUs were introduced initially in \citep{cho2014learning} and constitute a variant of the RNN architecture and LSTM especially. While LSTMs have two different states passed through the cells, cell state and hidden state, GRUs only contain one hidden state transferred between time steps. Furthermore, a GRU cell contains only two gates, update gate $u_{t}$ and reset gate $r_{t}$. 
The update gate determines the amount of information stored in the previous hidden state that would be retained for the future. It is quite similar to the input and forget gate in the LSTM. However, the control of new memory content added to the network is a new addition in GRUs. The reset gate is used from the model in order to decide how much of the past information to forget. Having a simpler architecture than LSTM, GRU requires less computation and can be trained faster. The basic components of a GRU cell are illustrated in Figure \ref{fig:gru_architecure} while the neural computations are:

\begin{equation}
u_{t} = \sigma(W_{u}(h_{t - 1}, x_{t}) + b_{u}) 
\end{equation}
\begin{equation}
r_{t} = \sigma(W_{r}(h_{t - 1}, x_{t}) + b_{r})
\end{equation}
\begin{equation}
\hat{h}_{t} = \varphi(W_{h}(r_{t}\odot h_{t-1}, x_{t}) + b_{h})
\end{equation}
\begin{equation}
h_{t} = u_{t}\odot h_{t-1} +(1 - u_{t}) \odot \hat{h}_{t}
\end{equation}

where $W_{u}$, $W_{r}$ and $W_{h}$ are weight matrices for the corresponding inputs of the network activation functions; $x_{t}$ is the input vector; $h_{t - 1}$ is the previous hidden state; $\hat{h}_{t}$ is the current memory function; $b_{u}$, $b_{r}$, $b_{h}$ are the corresponding bias vectors; $\odot$ is the Hadamard product; $\sigma$ is generally a sigmoid activation and $\varphi$ represents a non-linear function.

GRUs perform comparably to LSTMs, however as GRUs contain fewer parameters
they train generally faster due to the lighter computation. 

\subsection{NeuralProphet}

NeuralProphet is a newly developed Neural Network based PyTorch implementation and is heavily inspired by FBProphet (Section \ref{fbprophet}). NeuralProphet is a time series forecasting model consisting of trend, seasonality, auto-regression, special events, future regressors and lagged regressors. Specifically, trend is modelled either as a linear or a piece-wise linear trend with the use of changepoints. Seasonality is modelled using Fourier terms making the model able to handle multiple seasonalities for high-frequency data. An implementation of AR-Net \citep{DBLP:journals/corr/abs-1911-12436}, an Auto-Regressive Feed-Forward Neural Network for time series is exploited to handle auto-regression. Future regressors are external variables with known future values for the forecast period while the lagged regressors are external variables that only have values for the observed period. Lagged regressors are also modelled by exploiting a separate Feed-Forward Neural Network similar to the auto-regression. Finally, future regressors and special events are both modelled as covariates of the model with dedicated coefficients. More details on NeuralProphet can be found in the webpage of the project\footnote{\url{http://neuralprophet.com/}} and its git repository\footnote{\url{https://github.com/ourownstory/neural_prophet}}. Because NeuralProphet has a high number of hyperparameters, the default values for these were used in the experimental evaluation that follows.

\subsection{Time series forecasting after clustering}
\label{clustering}

To further exploit the neural networks introduced in the two previous sections, a clustering approach is first employed on the time series and then, a neural network model is deployed for each cluster. The main idea behind this approach is that neural networks require large amounts of data for training. Therefore, multiple time series that share the same characteristics and are in the same cluster would yield adequate samples for the training of the neural networks. Several approaches have been proposed in the literature for time series clustering, ranging from density-based methods \citep{DBLP:journals/itiis/AhnL18,DBLP:journals/information/HuangBZF19,DBLP:journals/jiis/LaurinecLLR19} to k-centroid methods \citep{DBLP:journals/tods/PaparrizosG17,DBLP:journals/eswa/RuizJAM20}. Although both approaches require a suitable set of time series features to properly identify clusters, the latter approaches further require apriori knowledge of the number of clusters or centroids $k$ which is not always available. To this end, a modified version of the famous DBSCAN \citep{Ester:1996:DAD:3001460.3001507} algorithm was employed which is a density-based clustering algorithm. Our proposed DBSCAN version uses two parameters to specify the proximity of candidate time series, namely the cosine similarity $cos(\theta)$ and the euclidean distance $\varepsilon$.

\textbf{Cosine similarity.} The cosine similarity is a measure that defines the angular distance between two non-zero vectors. Let $\Vec{u},\Vec{v}$ be two equally sized, non-zero vectors or time series, then the cosine similarity between them is:
\begin{equation} \label{cosine}
    cos(\theta) = \frac{u \cdot v}{||u|| \times ||v||} = \frac{\sum_{i=1}^n u_i \times v_i}{\sqrt{\sum_{i=1}^n u_i^2} \times \sqrt{\sum_{i=1}^n v_i^2}}
\end{equation}

where $n$ is the number of data points of the time series. Specifically, two vectors or time series with the same orientation yield a similarity of $1$, while two time series that are diametrically opposed yield a similarity of $-1$. Finally, two time series oriented at $90^{\circ}$ relative to each other yield a similarity of $0$.

\textbf{Euclidean distance.} Let $\Vec{u},\Vec{v}$ be two equally sized, non-zero vectors or time series, then the euclidean distance between them is:
\begin{equation} \label{euclidean}
    \varepsilon = || \Vec{u}-\Vec{v} || = \sqrt{(u_1 - v_1)^2 + (u_2 - v_2)^2 + ... + (u_n - v_n)^2}
\end{equation}

where $n$ is the number of data points of the time series. In order for the two time series to be in the same cluster both the cosine similarity and the euclidean distance thresholds must be met. Finally, after the clustering process is complete, multiple neural networks (either LSTM or GRUs as described in Sections \ref{LSTM} and \ref{GRUs} respectively) are deployed -- one for each cluster -- to train and test.

\section{Experimental Evaluation}
\label{evaluation}

In this section, we describe the dataset used for the experiments as well as the pre-processing steps followed to be suitable for experimentation. Furthermore, the forecasting capabilities of each algorithm described in Section \ref{models} are presented.

\subsection{Dataset description}
\label{dataset}

The dataset consists of $218,588$ water consumption time series from the region of Attica, Greece, spanning from 2013 to 2019 with quarterly frequency and were provided by the Water Supply and Sewerage Company of Greece\footnote{\url{https://www.eydap.gr/}}. Each time series is accompanied by a region id, denoting the city inside Attica it belongs to (e.g., Athens) and the water consumption is measured in cubic meters.

One challenge posed by the dataset is that the timestamp of each measurement or data point in a time series does not align with the timestamp of each measurement in other time series. For instance, the first measurement of one time series might be taken in January while the first measurement of another time series of the same quarter and year might be taken in March. To this end, we grouped time series together where each data point of the time series was taken in the same month. Furthermore, we grouped each time series by the region id. This resulted in multiple groups where each group was time aligned and from the same region.

\begin{figure}[!h]
\includegraphics[width=0.5\columnwidth]{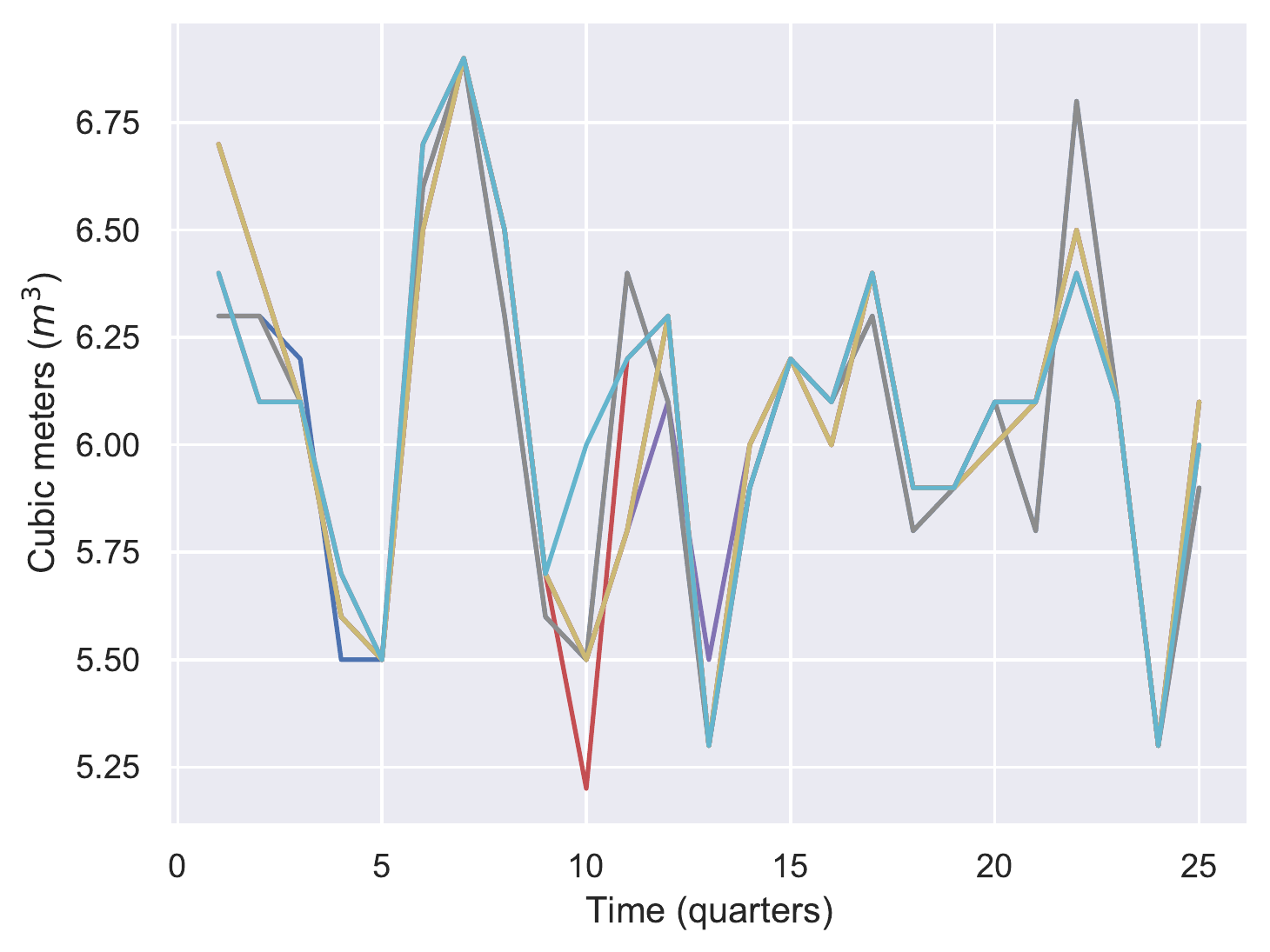}
\centering
\caption{Example of time series in the same cluster with $\varepsilon \leq 10$ and $cos(\theta) \geq 90\%$.}
\label{cluster_example}
\end{figure}

Another major challenge of the dataset is the fact that each time series does not belong to one user, but it might belong to a building or a group of buildings or a block, thus resulting in a large inhomogeneity. To tackle the inhomogeneity, the modified DBSCAN was applied in order to cluster similar time series together, thus capturing common seasonality patterns. The modified DBSCAN was applied in each group of time-aligned time series from the same region resulting in even more clusters. Finally, several parameters as described in Section \ref{clustering} of the DBSCAN were tested resulting in different datasets. Table \ref{tab:dataset} illustrates the parameters used along with the resulting number of time series and noise. Specifically, four datasets were resulted, D1 to D4, and they are presented from the more relaxed clustering parameters to the more strict ones. The minimum points ($minPts$) denote the least number of time series required to form a cluster, $\varepsilon$ denotes the euclidean distance threshold in cubic meters, whereas $cos(\theta)$ denotes the cosine similarity threshold in percentage. Finally, the last column indicates the percentage of the original dataset that was removed and considered to be outliers. We can observe that the more relaxed the parameters are, the smaller is the percentage of the dataset that is considered as noise and the more clusters are formed, which is to be expected. Figure \ref{cluster_example} illustrates time series that are in the same cluster. It can be seen that their observation values (measurements in cubic meters) do not deviate much from one another and that they all form common seasonality patterns. In the Section that follows, experiments were conducted on all of the resulting datasets.

\begin{table}[!h]
\centering
\begin{tabular}{|c|c|c|c|c|c|c|} 
 \hline
 Name & $minPts$ & $\varepsilon$ & $cos(\theta)$ & $\#$ time series & $\#$ of clusters & Noise \\
 \hline
 D1 & $10$ & $\leq 10$ & $\geq 80\%$ & $70,747$ & $348$ & $86.89\%$ \\ 
 D2 & $10$ & $\leq 10$ & $\geq 85\%$ & $46,421$ & $315$ & $90.78\%$ \\ 
 D3 & $10$ & $\leq 10$ & $\geq 90\%$ & $30,391$ & $267$ & $93.27\%$ \\ 
 D4 & $10$ & $\leq 5$ & $\geq 90\%$ & $21,491$ & $221$ & $95.34\%$ \\ 
 \hline
\end{tabular}
\caption{Datasets that resulted from the different DBSCAN parameters.}
\label{tab:dataset}
\end{table}

\subsection{Experimental Results}
\label{results}

To evaluate the effectiveness and the ability of its approach to forecast the time series, we split each time series into $80\%$ for training and $20\%$ for testing. In the case of neural networks after the clustering process, LSTM and GRU networks used $80\%$ of each time series in the same cluster for training, resulting in a forecasting model per cluster. Furthermore, each approach was evaluated to all four datasets as presented in Section \ref{dataset}, thus, acting as a means of evaluating the effect of the clustering process in the forecasting ability of LSTM and GRU networks and as a way to compare each methodology with one another. To properly evaluate the forecasts of each algorithm we used the following metrics:

\begin{itemize}
    \item Mean Absolute Error (MAE) is an average of the absolute errors $|e_i| = |y_i - x_i|$, where $y_i$ is the forecasted value and $x_i$ is the actual value.
    \item Mean Squared Error (MSE) is the average squared difference between the forecasted values and the actual values. This measure has the characteristic of penalizing errors due to the squared difference, thus acting as a measure to distinguish approaches with larger errors in their forecasts.
    \item Root Mean Squared Error (RMSE) is simply the square root of the MSE.
\end{itemize}

\begin{figure}[ht!]  
    \begin{subfigure}[t]{.50\textwidth}
        \centering
        \includegraphics[width=\textwidth, keepaspectratio]{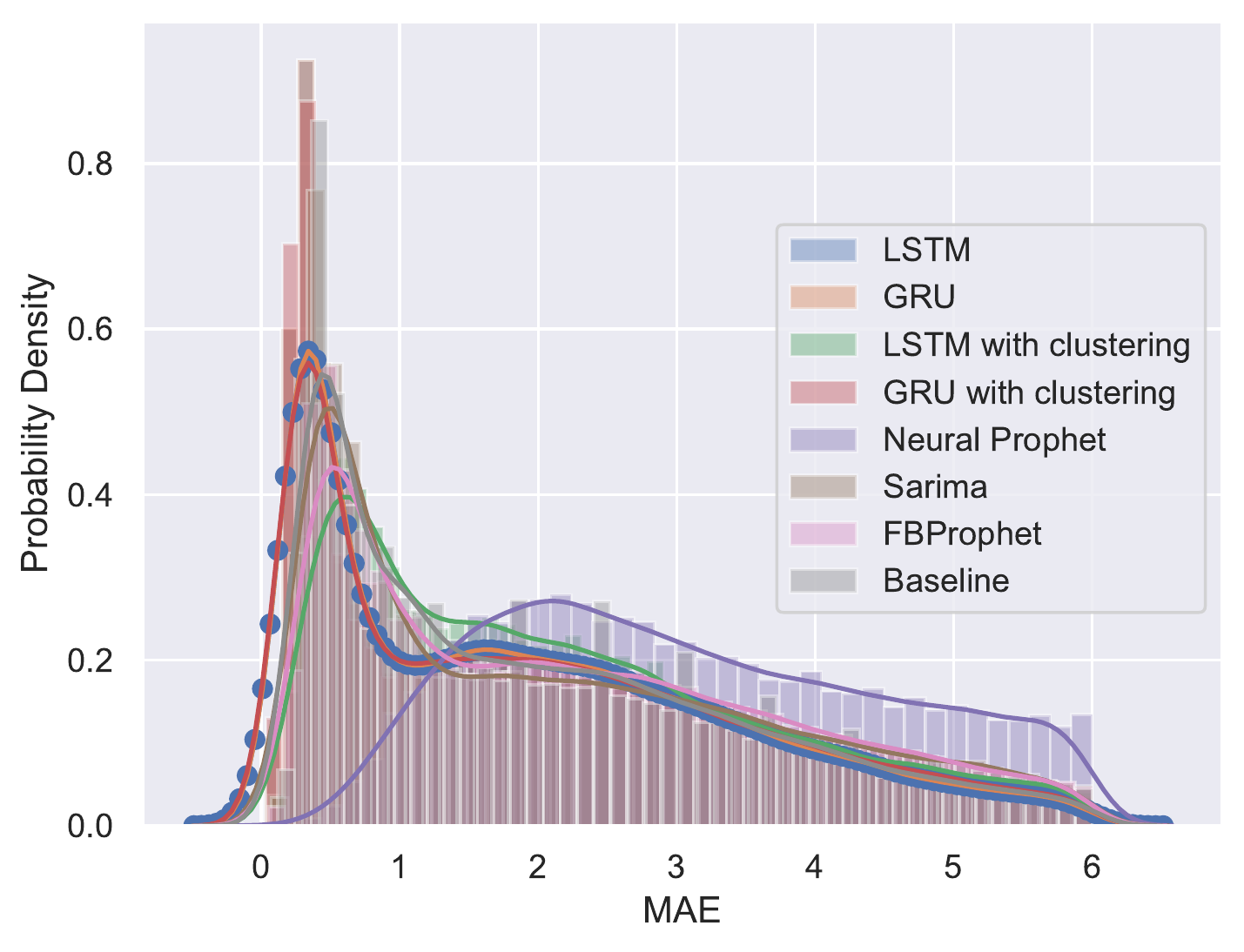}  
        \caption{Density Distribution of Mean Absolute Error}
        \label{mae_d1}
    \end{subfigure}\hfill
    \begin{subfigure}[t]{.50\textwidth}
        \centering
        \includegraphics[width=\textwidth, keepaspectratio]{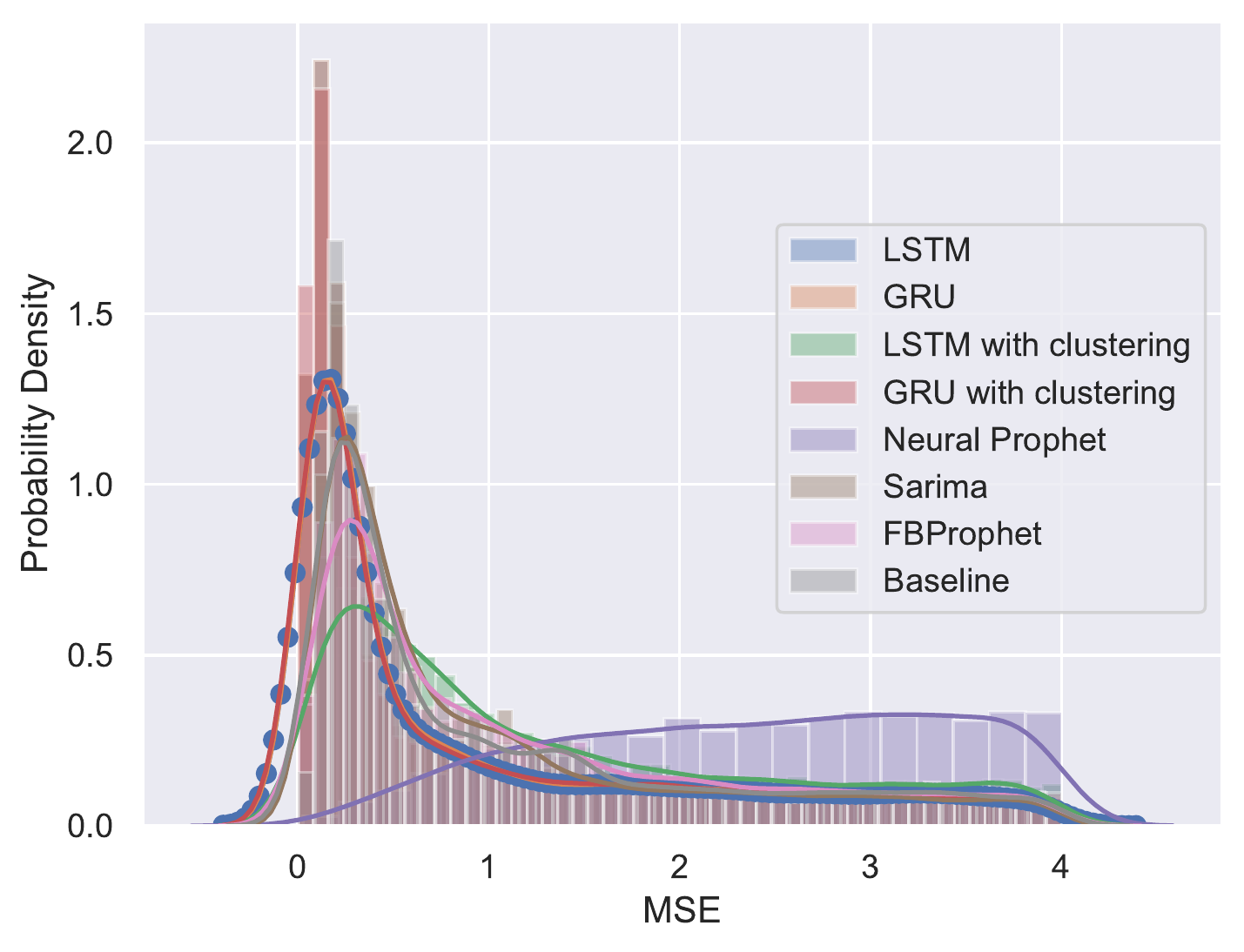} 
        \caption{Density Distribution of Mean Squared Error}
        \label{mse_d1}
    \end{subfigure}
\caption{Dataset D1}
\end{figure}  

\begin{figure}[ht!]  
    \begin{subfigure}[t]{.50\textwidth}
        \centering
        \includegraphics[width=\textwidth, keepaspectratio]{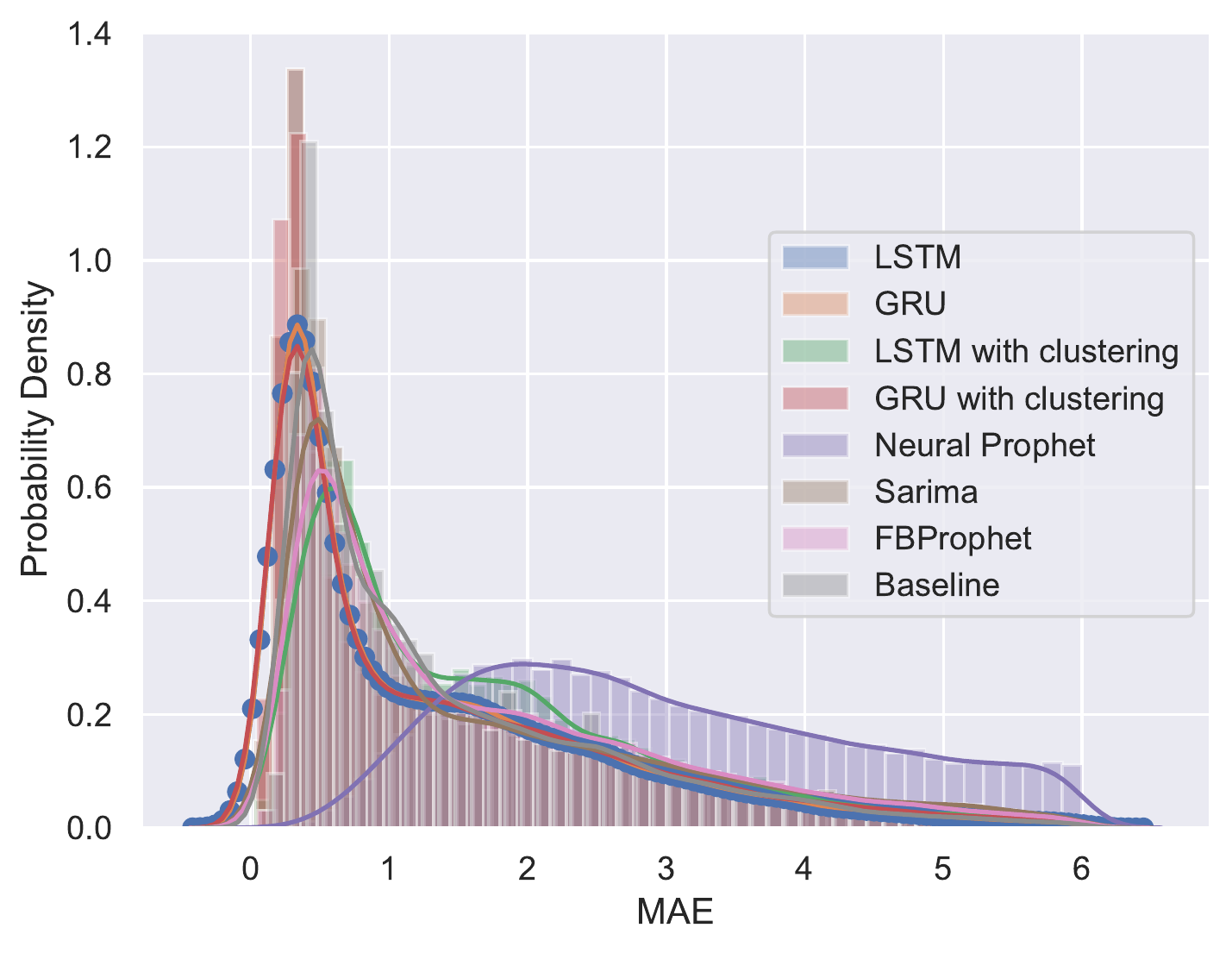}  
        \caption{Density Distribution of Mean Absolute Error}
        \label{mae_d2}
    \end{subfigure}\hfill
    \begin{subfigure}[t]{.50\textwidth}
        \centering
        \includegraphics[width=\textwidth, keepaspectratio]{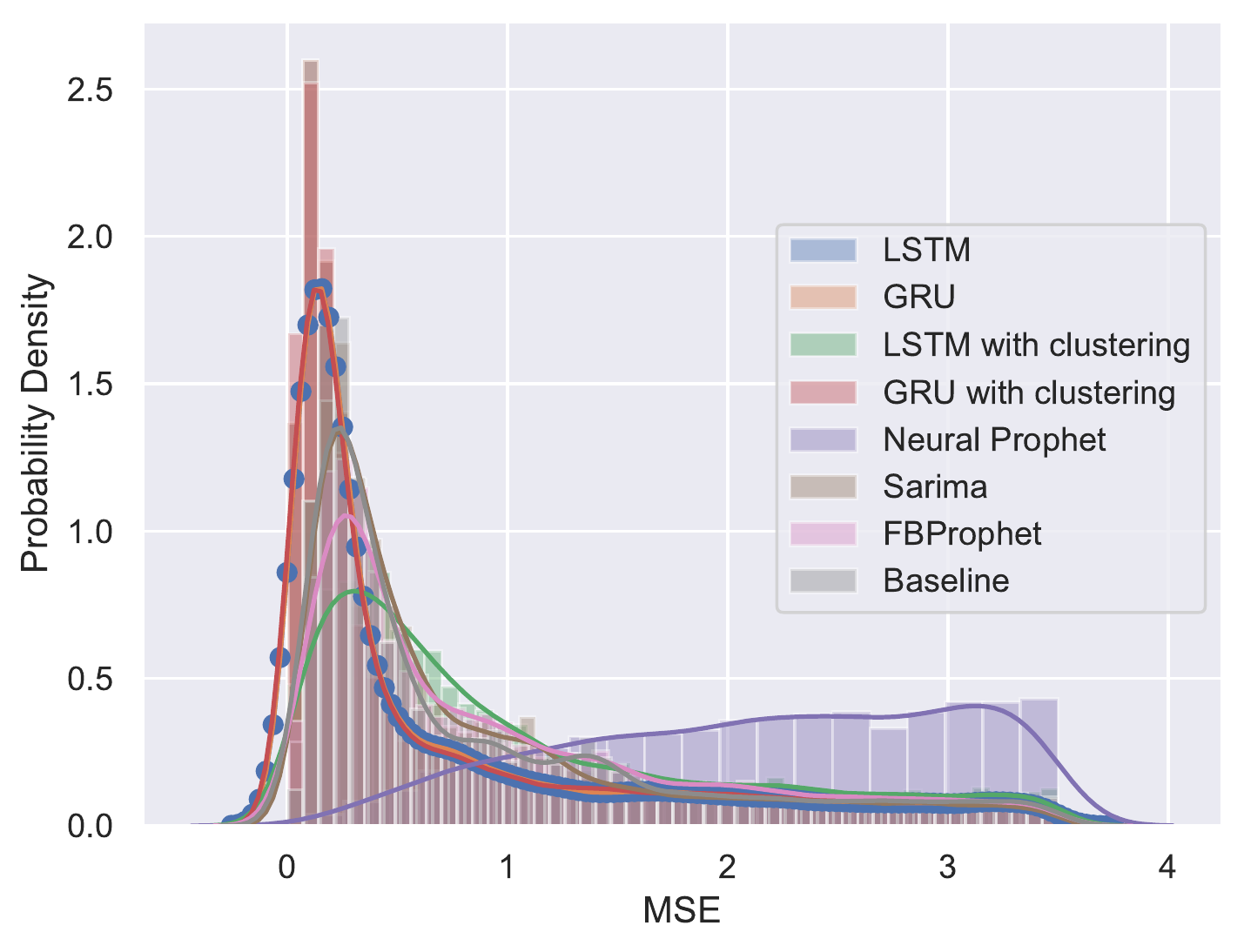} 
        \caption{Density Distribution of Mean Squared Error}
        \label{mse_d2}
    \end{subfigure}
\caption{Dataset D2}
\end{figure}

\begin{figure}[ht!]  
    \begin{subfigure}[t]{.50\textwidth}
        \centering
        \includegraphics[width=\textwidth, keepaspectratio]{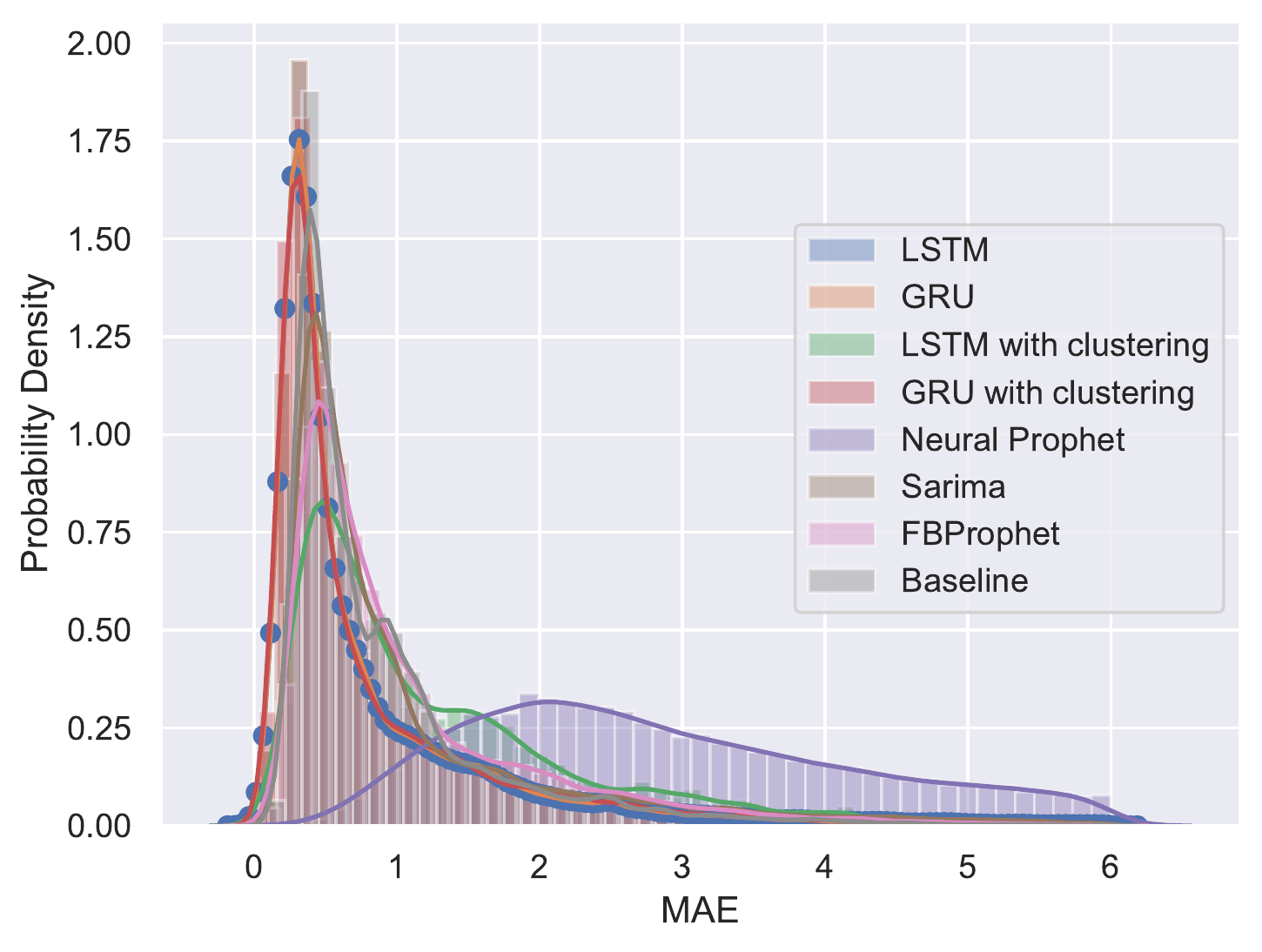}  
        \caption{Density Distribution of Mean Absolute Error}
        \label{mae_d3}
    \end{subfigure}\hfill
    \begin{subfigure}[t]{.50\textwidth}
        \centering
        \includegraphics[width=\textwidth, keepaspectratio]{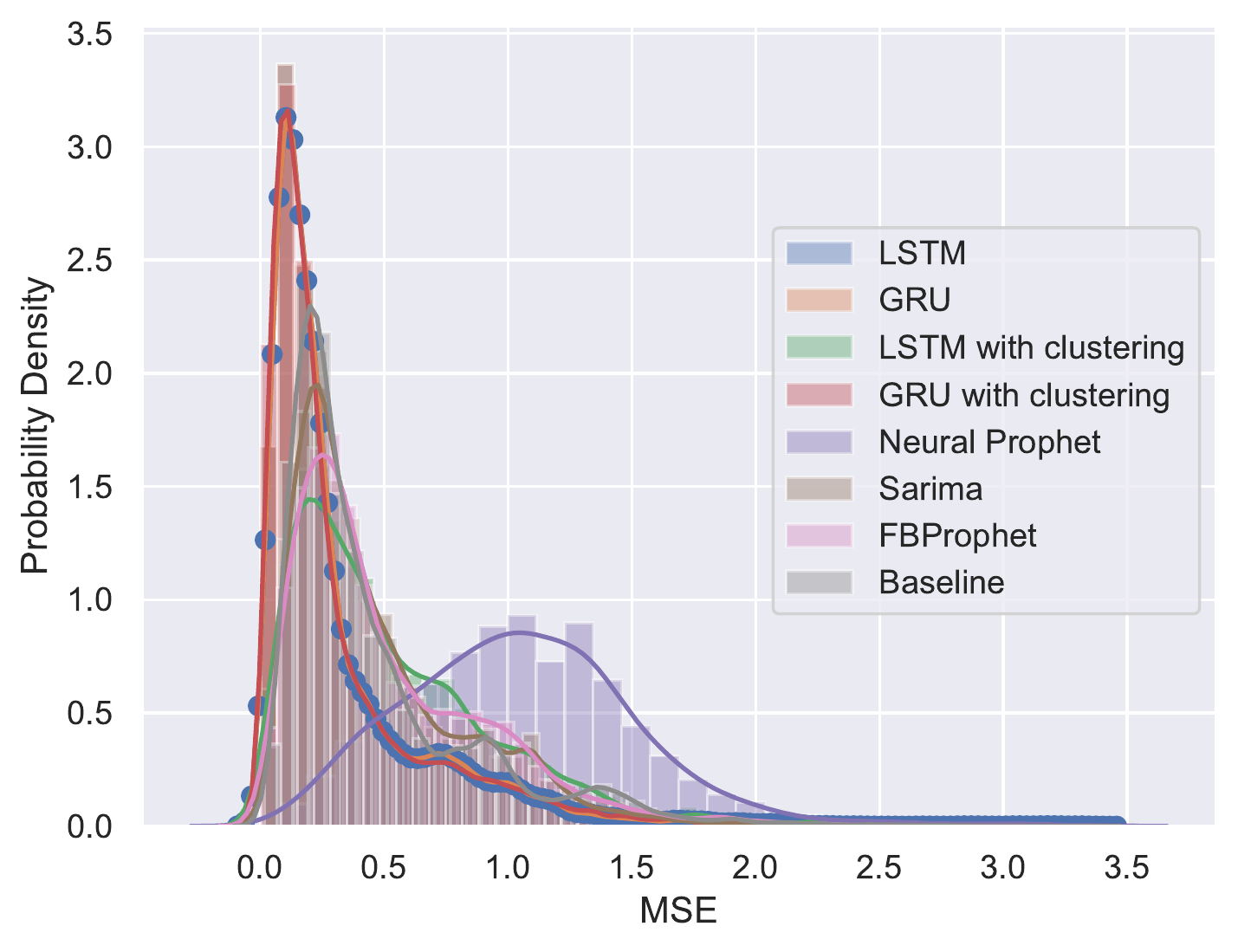} 
        \caption{Density Distribution of Mean Squared Error}
        \label{mse_d3}
    \end{subfigure}
\caption{Dataset D3}
\end{figure}

\begin{figure}[ht!]  
    \begin{subfigure}[t]{.50\textwidth}
        \centering
        \includegraphics[width=\textwidth, keepaspectratio]{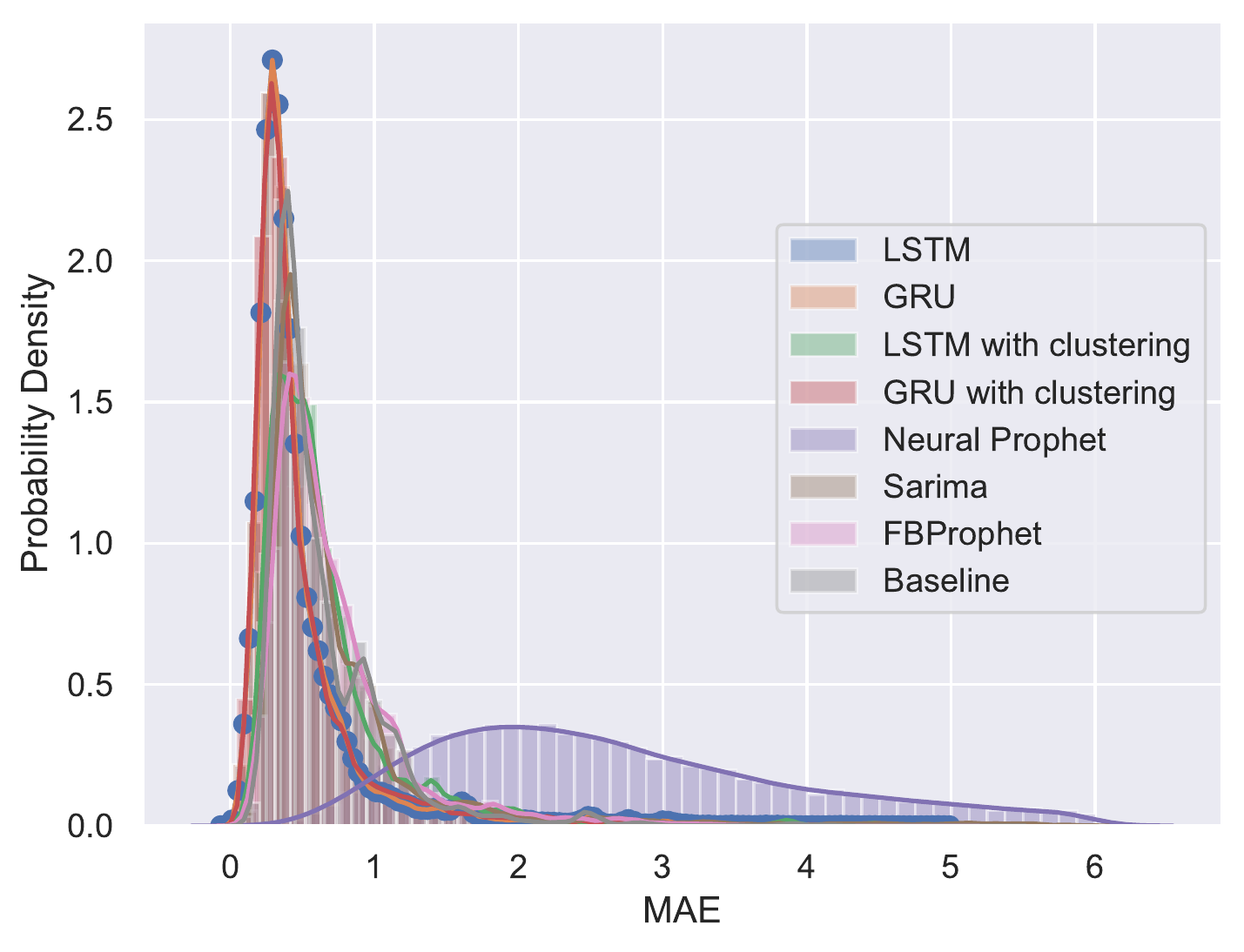}  
        \caption{Density Distribution of Mean Absolute Error}
        \label{mae_d4}
    \end{subfigure}\hfill
    \begin{subfigure}[t]{.50\textwidth}
        \centering
        \includegraphics[width=\textwidth, keepaspectratio]{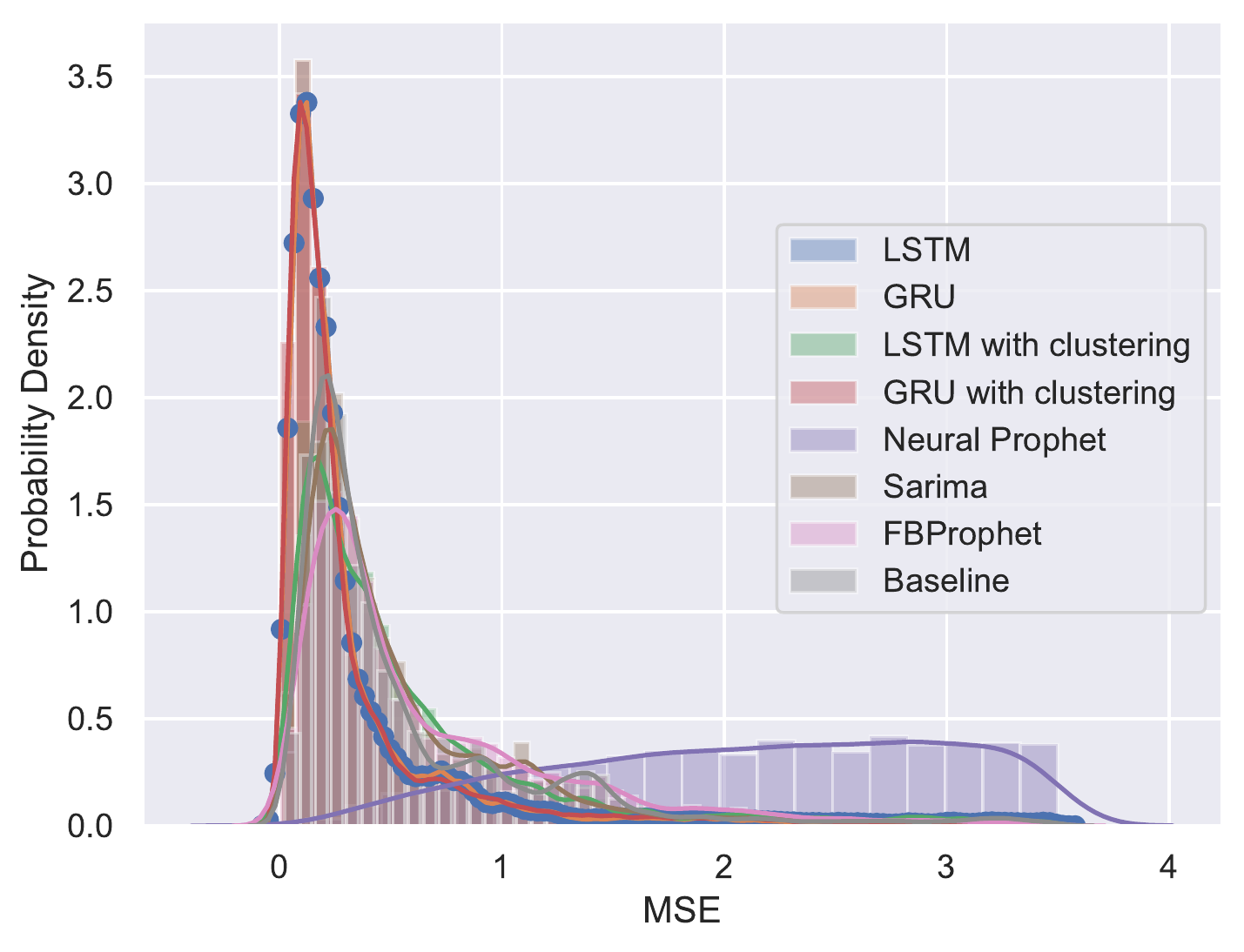} 
        \caption{Density Distribution of Mean Squared Error}
        \label{mse_d4}
    \end{subfigure}
\caption{Dataset D4}
\end{figure}

For the evaluation of the large number of time series in the datasets, we used a Kernel Density Estimation (KDE) figure of MAE, MSE and RMSE, which demonstrates the density of each evaluation measure in the entirety of the datasets. Figures \ref{mae_d1}, \ref{mae_d2}, \ref{mae_d3} and \ref{mae_d4} illustrate the MAEs for datasets D1, D2, D3 and D4 respectively. Similarly, Figures \ref{mse_d1}, \ref{mse_d2}, \ref{mse_d3} and \ref{mse_d4} illustrate the MSEs for datasets D1, D2, D3 and D4 respectively.



From the results, we can observe that LSTMs (blue circled line), GRUs (orange line) and GRUs with clustering (red line) demonstrate the lowest MAE and MSE values in the datasets, suggesting that these models are the most suitable choice for this type of data. Surprisingly, the baseline approach (grey line) follows the neural networks and SARIMA, FBProphet and LSTM with clustering are the next candidates after the baseline approach. This may be an indication that SARIMA and FBprohpet require more training data and that the sparsity of the time series (four observations per year) does not provide the volume of training required by these methodologies. Finally, Neural Prophet seems to be the least effective method to forecast the water consumption values. The reason for this behavior may be the fact that the default parameters of Neural Prophet were chosen and experimentation for the best parameters for this type of dataset was not conducted. The findings of this research agree with other papers in the literature that consider the neural networks to be the most suitable approach for time series forecasting \citep{xu2019hybrid,zheng2017electric,sagheer2019time}.

\begin{figure}[ht!]  
    \begin{subfigure}[t]{.50\textwidth}
        \centering
        \includegraphics[width=\textwidth, keepaspectratio]{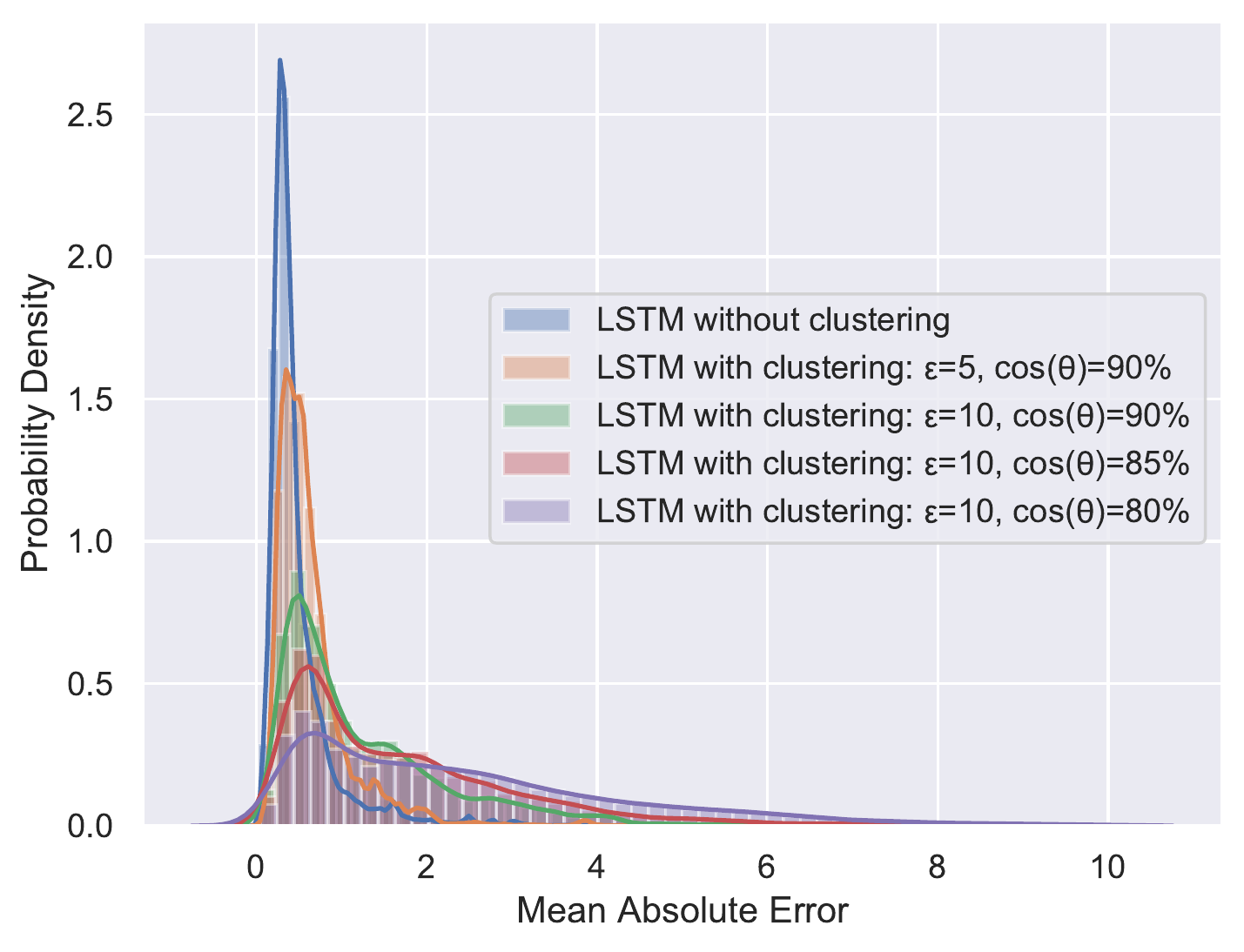}  
        \caption{LSTM}
        \label{users_mae_lstm}
    \end{subfigure}\hfill
    \begin{subfigure}[t]{.50\textwidth}
        \centering
        \includegraphics[width=\textwidth, keepaspectratio]{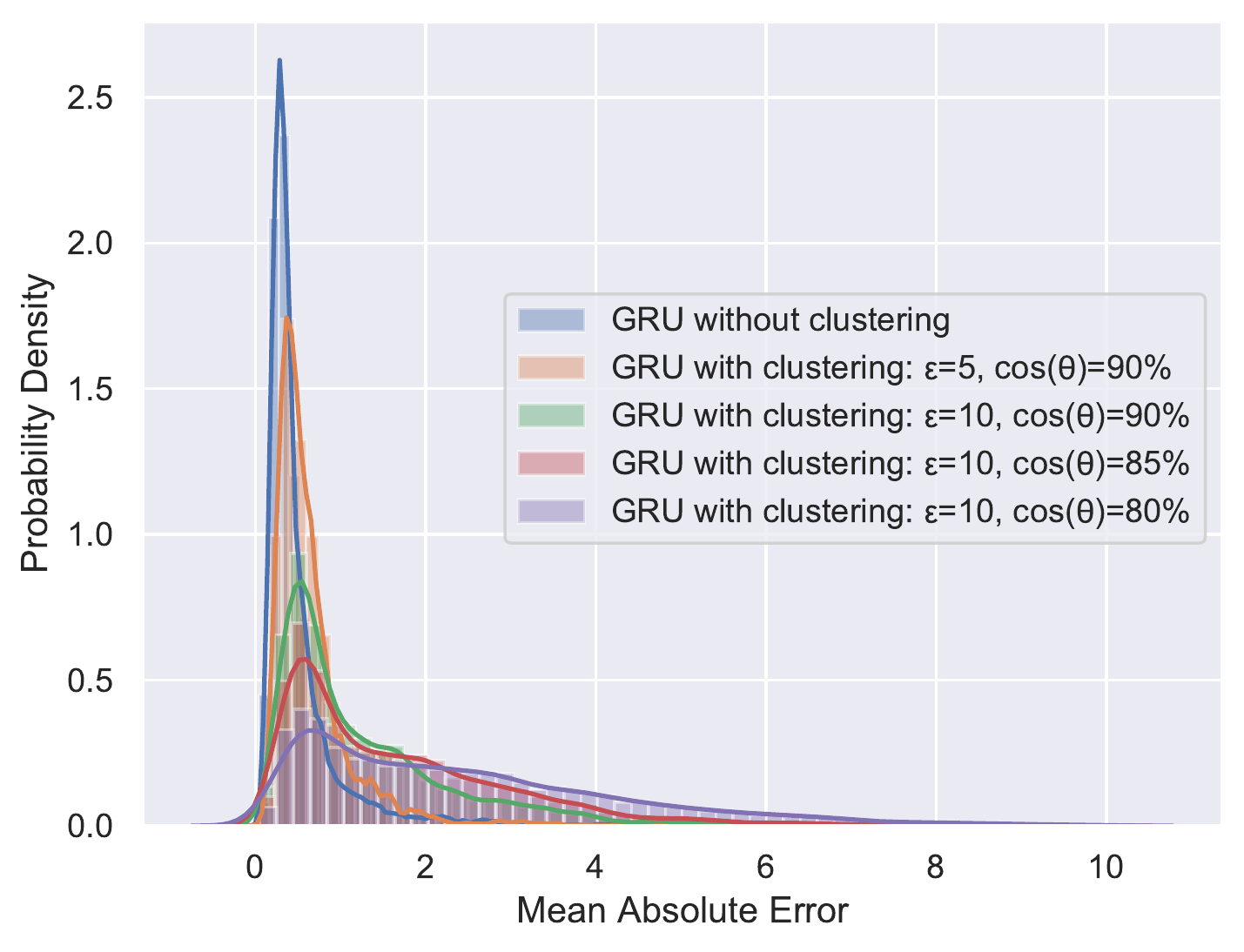} 
        \caption{GRU}
        \label{users_mae_gru}
    \end{subfigure}
\caption{Density Distribution of Mean Absolute Error}
\end{figure}



\begin{figure}[ht!]  
    \begin{subfigure}[t]{.50\textwidth}
        \centering
        \includegraphics[width=\textwidth, keepaspectratio]{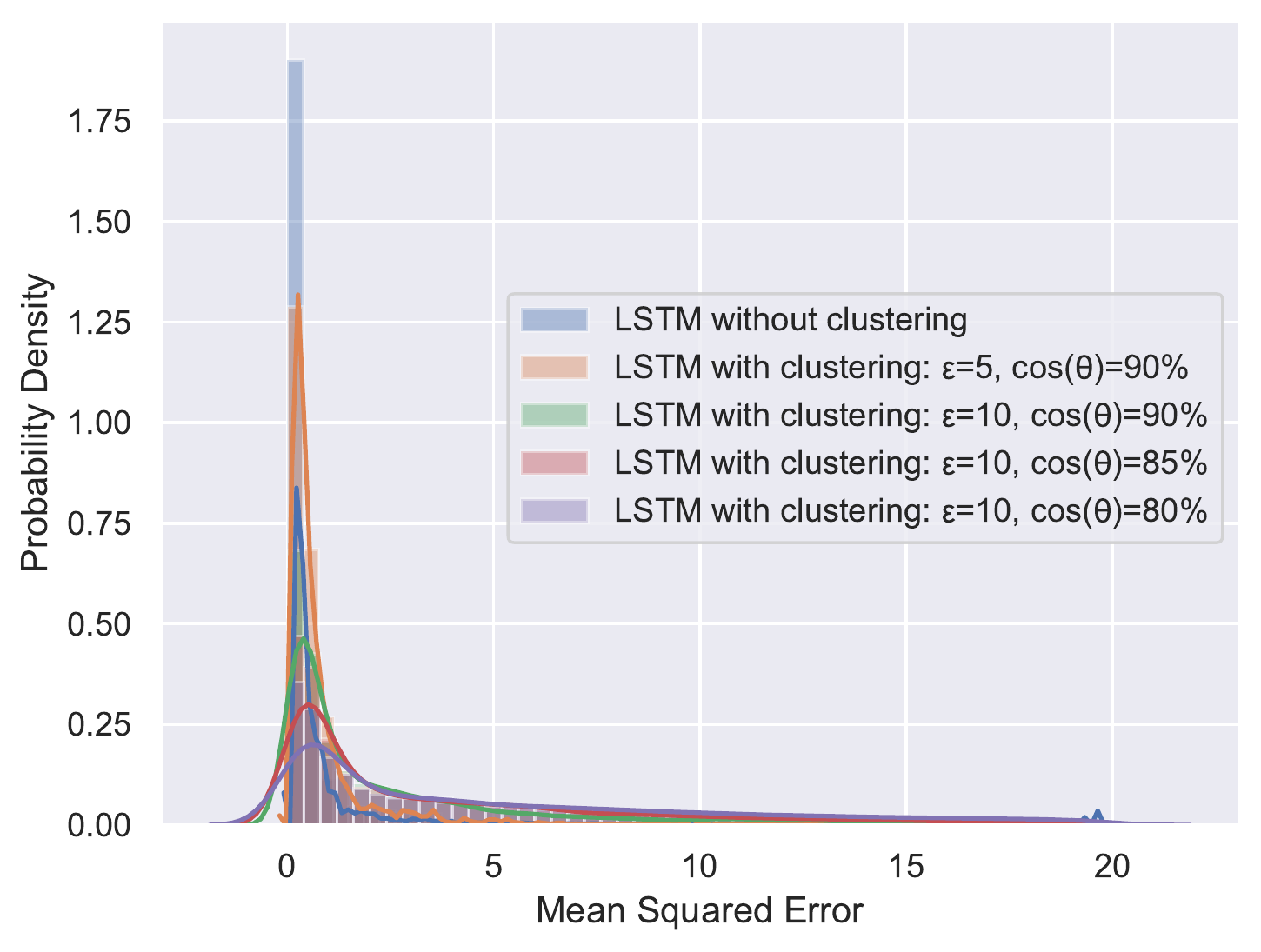}  
        \caption{LSTM}
        \label{users_mse_lstm}
    \end{subfigure}\hfill
    \begin{subfigure}[t]{.50\textwidth}
        \centering
        \includegraphics[width=\textwidth, keepaspectratio]{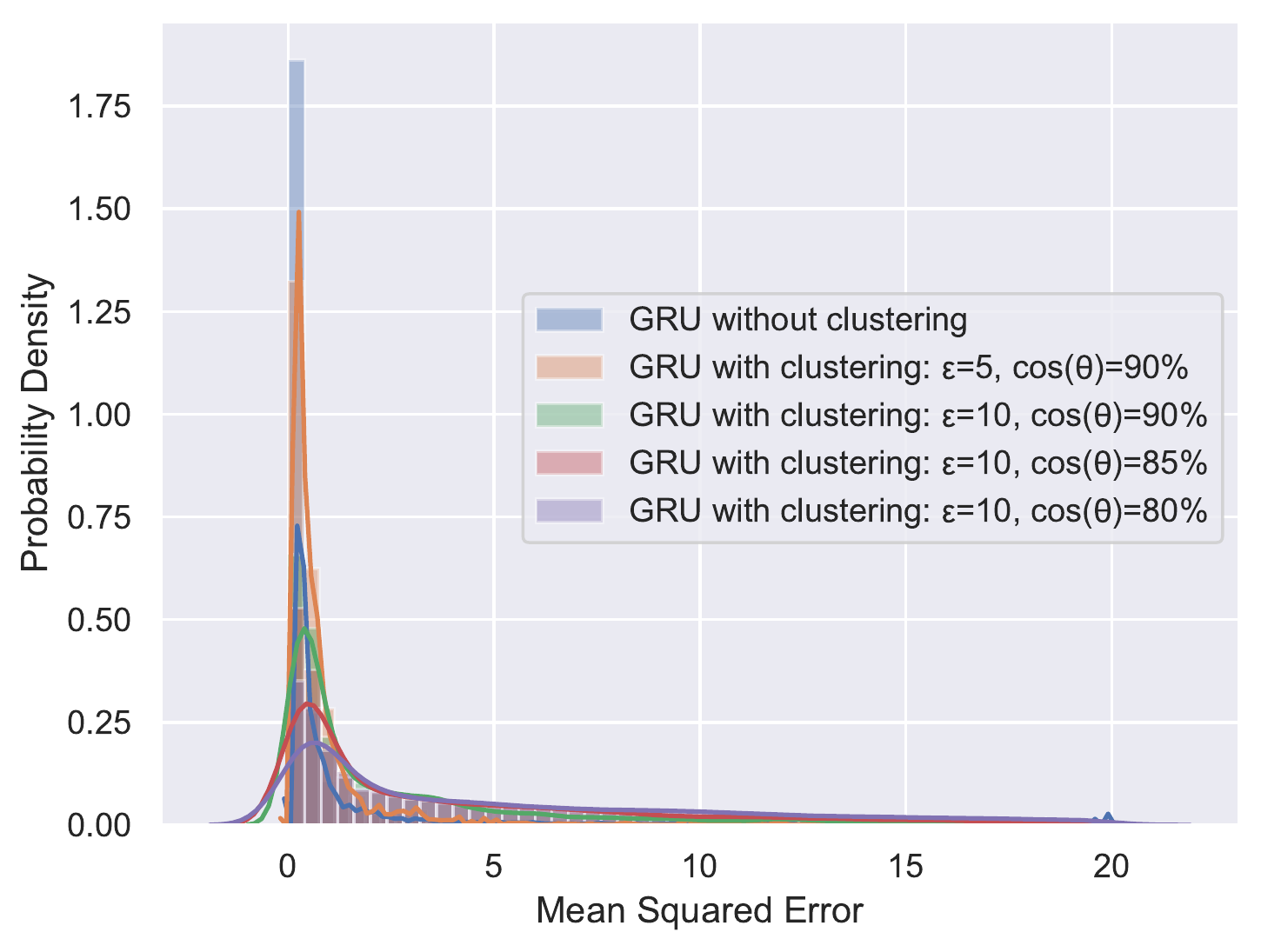} 
        \caption{GRU}
        \label{users_mse_gru}
    \end{subfigure}
\caption{Density Distribution of Mean Squared Error}
\end{figure}



\begin{figure}[ht!]  
    \begin{subfigure}[t]{.50\textwidth}
        \centering
        \includegraphics[width=\textwidth, keepaspectratio]{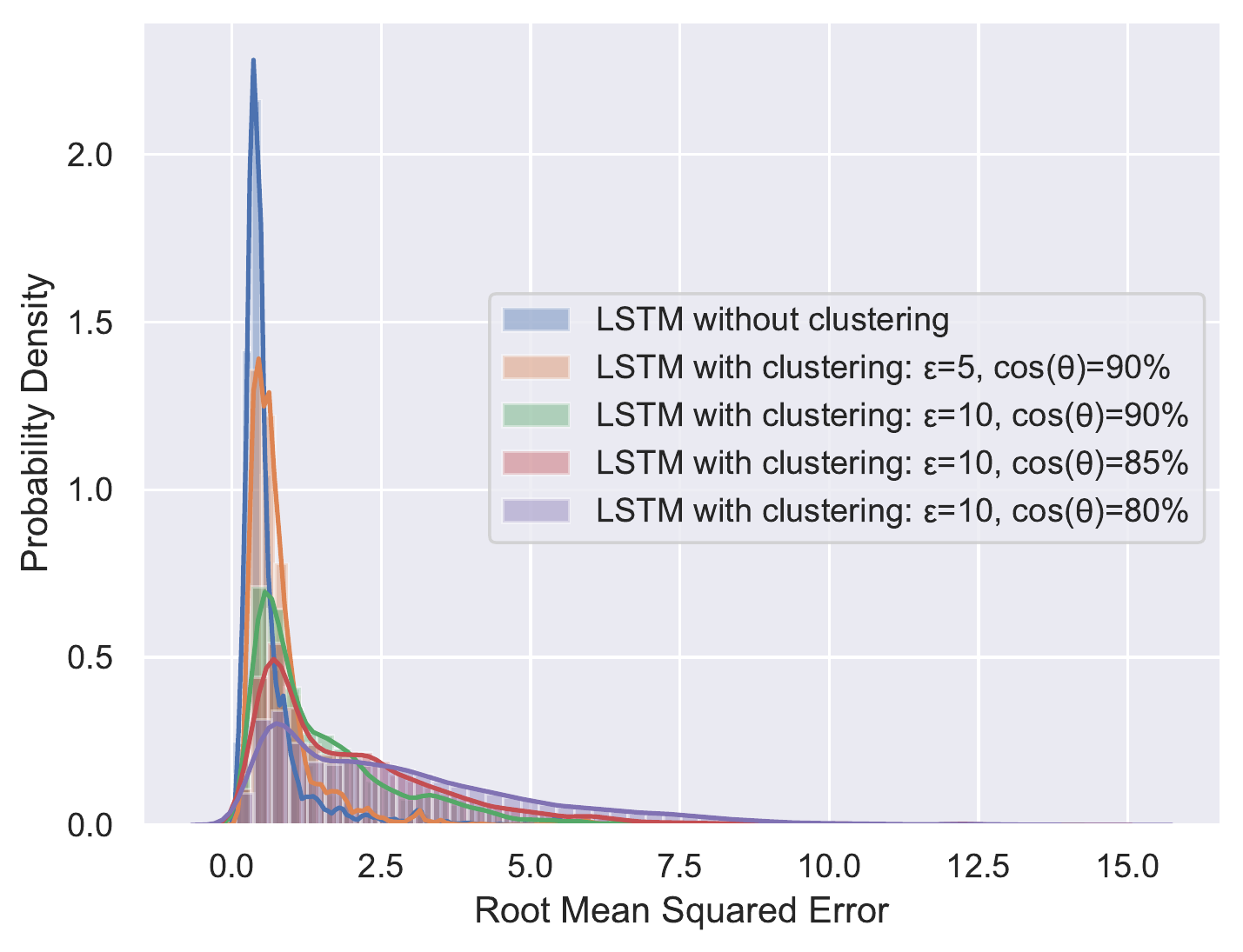}  
        \caption{LSTM}
        \label{users_rmse_lstm}
    \end{subfigure}\hfill
    \begin{subfigure}[t]{.50\textwidth}
        \centering
        \includegraphics[width=\textwidth, keepaspectratio]{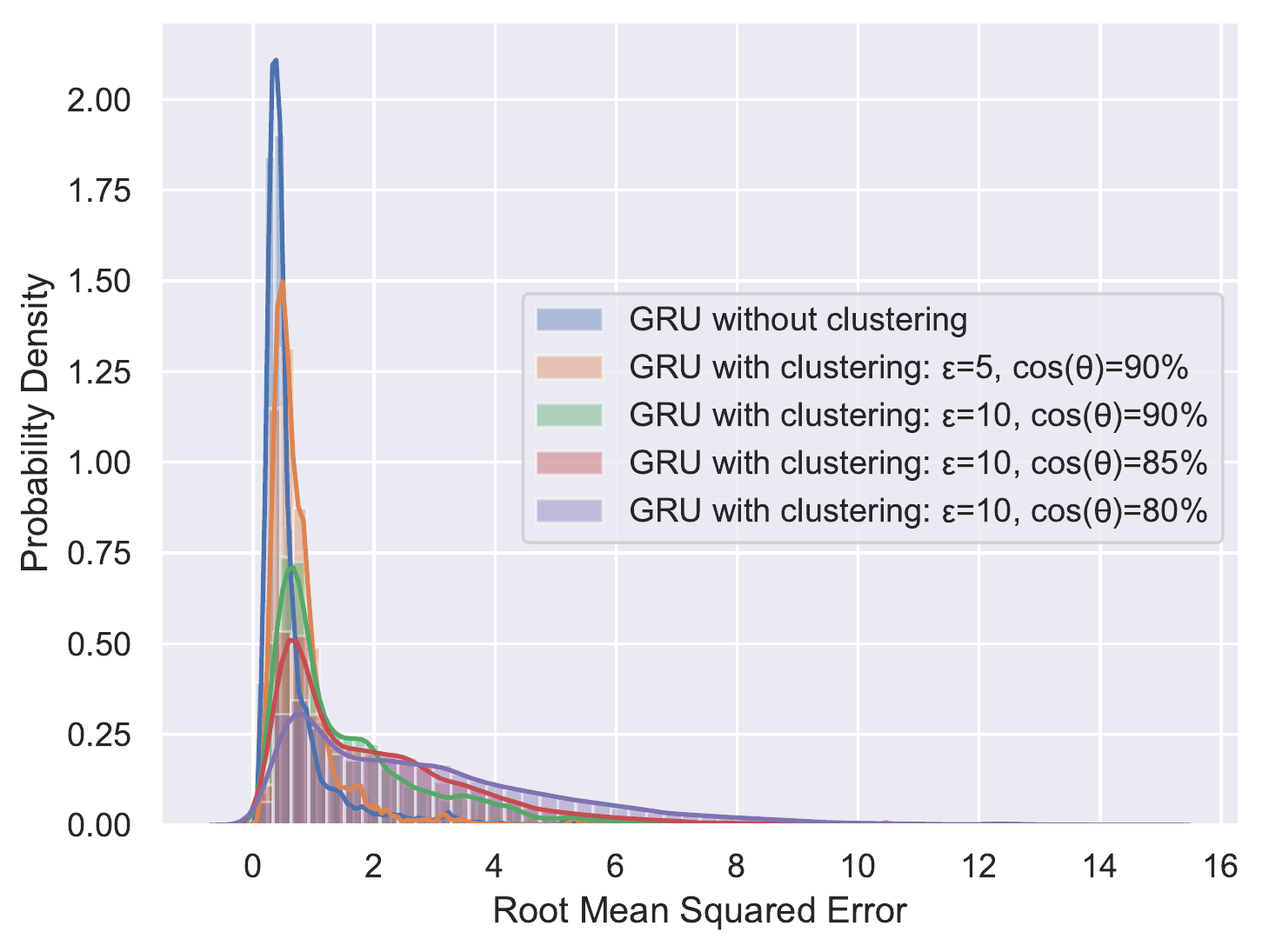} 
        \caption{GRU}
        \label{users_rmse_gru}
    \end{subfigure}
\caption{Density Distribution of Root Mean Squared Error}
\end{figure}  



To further investigate the effect of time series clustering in the forecasting ability of the neural networks, we visualized the density of the MAE, MSE and RMSE values in a KDE figure for each neural network methodology (LSTM, GRU), with and without the clustering of the series. Figures \ref{users_mae_lstm}, \ref{users_mse_lstm} and \ref{users_rmse_lstm} demonstrate the MAE, MSE and RMSE of the LSTM neural networks respectively. Similarly, Figures \ref{users_mae_gru}, \ref{users_mse_gru} and \ref{users_rmse_gru} illustrate the same evaluation metrics for the GRU neural networks. From these figures, we can observe that as the clusters become less relaxed and more dense the forecasting ability of the models improve. This is to be expected since time series that are closer to each other, in terms of observation values and seasonality, will have similar behavior in the future. Furthermore, despite the fact that LSTM and GRU neural networks after the clustering had more training data, experiments indicate that one model per time series is still more ideal for the forecasting of water consumption values. This is due to the fact that when more than one time series are used for the training of the neural networks, the observation values of all the series are averaged out, thus creating an overall average time series per cluster. Finally, it can be observed that the LSTM neural networks without the clustering perform better than the respective GRU neural networks and that the LSTM neural networks with clustering perform worse thatn the respective GRU neural networks. This can be explained by the fact that in the former case, each neural network model has only a few training samples, a case in which the LSTM neural networks can yield better forecasting performance. On the other hand, in the latter case, where more training data are available, we can observe that the GRU models outperform the LSTM models, making the GRU neural networks more suitable for forecasting time series with a larger amount of observation values.

\section{Conclusions}
\label{conslusion}
In this paper, several well-known time series forecasting algorithms were evaluated on a dataset provided by the Water Supply and Sewerage Company of Greece. The dataset provided posed several challenges such as time mis-alignments, sparse observation values of the time series and water consumption measurements that do not originate from a single user. The experimental results indicated that the neural networks had the best forecasting performance with statistical methodologies such as SARIMA and FBprophet not falling further behind. Furthermore, a clustering approach was evaluated in which a modified DBSCAN algorithm was applied first and then a neural network was trained over all time series in a cluster, resulting in a neural network per cluster. Experimental results demonstrated that the clustering approach did not outperform other methodologies in the study due to the fact that an average time series out of all the time series per cluster was learned from the neural networks. The aim of this study was to evaluate several well-known time series forecasting algorithms in a dataset with distinct challenges and provide insights for researchers in the field of time series forecasting.

\bibliographystyle{unsrtnat}


\end{document}